\documentclass{llncs}
\usepackage[authoryear]{natbib}
\usepackage{amsmath, amssymb}
\usepackage{tikz}

\title{Note on Representing attribute reduction and concepts in concepts lattice using graphs}
\author{Jan Konecny}

\institute{Dept.\,Computer Science, Palacky University, Olomouc\\ 17.\,listopadu 12, CZ-77146 Olomouc, Czech Republic}

\newcommand{\up}{{\uparrow}}
\newcommand{\down}{{\downarrow}}
\newcommand{\B}{\ensuremath{{\cal B}}}
\def\tu#1{(#1)}
\newcommand{\noticka}[1]{}
\def\X{\ensuremath{\!\times\!}}
\begin{document}

\maketitle{}

\begin{abstract}
Mao H. (2017, Representing attribute reduction and concepts in concept lattice using graphs.
{\em Soft Computing} 21(24):7293--7311)
claims to make contributions to the study of reduction of attributes in concept lattices by using graph theory.
We show that her results are either trivial or already well-known
and all three algorithms proposed in the paper are incorrect.
\end{abstract}


\section{Introduction}

\cite{Mao2016} claims that she studies attribute reduction 
and formal concept enumeration in Formal Concept Analysis (FCA) with the aid of Graph Theory. We show, that the use of the Graph Theory is trivial.
Specifically, Mao simply replaces the notion of formal context by an equivalent notion of pre-weighted relevant graph and presents 
well-known and trivial facts
as novel and interesting.
Additionally, Mao provides a method of enumeration
of formal concepts in a reduced formal context.
The method is represented by three algorithms;
we demonstrate, using examples, that they are  incorrect.

\section{Preliminaries}
We use the same notions and the same notations as \cite{Mao2016}; however we need to also recall a few notions from \citep{GaWi:FCA} to explain our case.
Thus, for the reader's convenience, we provide full preliminaries.

An input to FCA is a triplet $(O,P,I)$,
called a \emph{formal context}, where $O,P$ are finite non-empty sets of objects and attributes, respectively, and $I$ is a binary relation between $O$ and $P$; $\tu{o,a}\in I$ means that the object $o$ has the attribute $a$.
 Finite formal contexts are usually depicted as tables, in which rows represent objects, columns represent attributes, and each entry contains a cross if the corresponding object has the corresponding attribute, and is otherwise left blank 
 (see top parts of Figures~\ref{fig:cex1}--\ref{fig:cex3} for examples).

The formal context induces the following operators:
\begin{itemize}
\item[]
$\up: \mathbf{2}^O \to \mathbf{2}^P$ assigns to a set $X$ of objects the set $X^\up$ of all attributes shared by all the objects in $X$.
\item[]
$\down:\mathbf{2}^P \to \mathbf{2}^O$ assigns to a set $B$ of attributes the set $B^\down$ of all objects which share all the attributes in $B$.
\end{itemize}
For singletons we use shortened notation and write $o^\up$, $a^\down$ instead of $\{o\}^\up$, $\{a\}^\down$, respectively.

{\em Formal concept} is a pair $\tu{X,B}$ of sets $X \subseteq O, B \subseteq P$, s.t. $X^\up = B$ and $B^\down = X$.
The first component of a formal concept is called extent, the second one is called intent.
The collection of all formal concepts in $(O,P,I)$ is denoted $\B(O,P,I)$.
The collection $\B(O,P,I)$ with order $\le$ defined by 
$\tu{X_1,B_1} \le \tu{X_2,B_2}$ if $X_1 \subseteq X_2$ 
for all formal concepts $\tu{X_1,B_1}, \tu{X_2,B_2} \in \B(O,P,I)$,
forms a complete lattice called a {\em concept lattice}.

We call a computation of $\B(O,P,I)$ a {\em formal concept enumeration}.

We consider the following two binary relations on $P$ induced by a formal context:
\begin{itemize}
\item dependency $\sqsubseteq$, defined by
$a \sqsubseteq b$ if $a^\down \subseteq b^\down$
for all $a,b\in P$,
\item equivalence $\equiv$, defined by
$a \equiv b$ if $a^\down = b^\down$
for all $a,b\in P$.
\end{itemize}

Formal context $(O,P,I)$ is called 
{\em clarified} if
$a_1 \equiv a_2$ implies $a_1 = a_2$
for any $a_1,a_2\in P$; i.e. if it does not contain duplicate columns.
Removal of duplicate columns (keeping one representative of each $\equiv$-class) is called a {\em clarification}.

An attribute $a$ is called {\em reducible} in formal context $(O,P,I)$
if there is $Z\subseteq P\setminus\{a\}$ such that $a^\down = Z^\down$
(equivalently, if $\B(O,P,I)$ and $\B(O,P\setminus \{a\},I\cap (O \times P\setminus \{a\}))$
are isomorphic).
Formal context is {\em  reduced} if it has no reducible attributes. Removal of reducible 
attributes is called {\em a reduction}.
\cite{GaWi:FCA} provide efficient method for
clarification and reduction.

Originally for Rough Set Theory,
\cite{Pawlak} proposed three types of attributes.
These types were introduced into FCA by \cite{Zhang2005}. We omit their definitions and just explain their relationship
to the reducible and irreducible attributes: {\em absolutely necessary (core) attributes} are exactly irreducible attributes;
{\em relatively necessary attributes} are those reducible attributes that become irreducible when their duplicates are removed;
and {\em absolutely unnecessary attributes} are those which are neither absolutely nor relatively necessary.

\section{The use of Graph Theory is trivial and the results are not novel}
\label{sec:red}

 \cite{Mao2016} claims to propose a method of attribute reduction
which utilizes Graph Theory and is 
based on removing vertices from a particular graph; this should distinguish her
method from other known approaches \citep{GaWi:FCA,Zhang2005} which are based on
the removal of attributes.

Specifically, she introduces a so-called
{\em pre-weighted relevant graph} $G(O,P,I )$ (derived from a formal context $(O,P,I)$) as 
a graph with vertices being the attributes in $P$ and
 edges being given as follows: for $a, b \in P$ and $a \neq b$,
 \begin{itemize}
 \item
 if $x^\down =y^\down$  there is a bi-arc joining $x$ and $y$, i.e. $x \Leftrightarrow y$,
 \item
 if $x^\down \subset y^\down$ there is an arc joining $y$ to $x$, i.e. $x \to y$.
 \end{itemize}

Thus, $G(O,P,I)$ is basically 
$\tu{O,P,I}$ with the relations $\sqsubseteq$ and $\equiv$ (specifically, the arcs $\to$ correspond to pairs in $\sqsubseteq \setminus \equiv$ 
and the bi-arcs $\Leftrightarrow$ correspond to non reflexive pairs in $\equiv$).
This represents the entire utilization of Graph Theory in  Mao's work. 

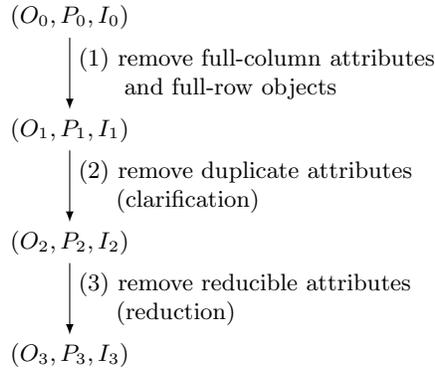
\begin{figure}
\begin{center}
\begin{tikzpicture}
\node (0) at (0,0) {$(O_0,P_0,I_0)$};
\node (1) at (0,-1.5) {$(O_1,P_1,I_1)$};
\node (2) at (0,-3) {$(O_2,P_2,I_2)$};
\node (3) at (0,-4.5) {$(O_3,P_3,I_3)$};

\draw[-latex] (0) -- node[right,rectangle, text width=5cm] {(1) remove full-column attributes\\\hspace{.5cm} and full-row objects}  (1);
\draw[-latex] (1) -- node[right,rectangle, text width=5cm] {(2) remove duplicate attributes \\\hspace{.5cm}(clarification)}(2);
\draw[-latex] (2) -- node[right,rectangle, text width=5cm] {(3) remove reducible attributes \\\hspace{.5cm}(reduction)}(3);
\end{tikzpicture}
\caption{Three phase process of attribute reduction in \cite{Mao2016}\label{fig:process}.}
\end{center}
\end{figure}

Mao describes the attribute reduction in three steps (see Fig.\,\ref{fig:process}):
\begin{itemize}
\item[(1)] For an input formal context $(O_0,P_0,I_0)$ satisfying
\begin{equation*}
o^\up \neq \emptyset\text{ and } a^\down \neq \emptyset \text{ for all $o\in O_0, a\in P_0$}
\end{equation*}
the formal context $(O_1,P_1,I_1)$ is found by removing full-row objects and full-column attributes from $(O_0,P_0,I_0)$.
\item[(2)] For $(O_1,P_1,I_1)$ 
find context $(O_2,P_2,I_2)$
by clarification.
\item[(3)] For $(O_2,P_2,I_2)$ find context $(O_3,P_3,I_3)$
by reduction.
\end{itemize}

The method is basically identical with the method known from the basic literature \citep{GaWi:FCA}.

At best, Mao's results could be considered to show how to use precomputed
relations $\sqsubseteq$ and $\equiv$ to achieve better time complexity.
However there are two problems with it.
\begin{itemize}
   \item First, the relations are used the same way 
   as in \citep{GaWi:FCA} (where they are not precomputed), or they 
   present a trivial improvement. As an example of the latter,
   \citep[Theorem 3.1]{Mao2016} 
   characterizes 
   objects with full row.
   The theorem states  that we need not check whether the object
   has all attributes, instead we can just check whether it has all attributes
   that are minimal w.r.t. $\sqsubseteq$.
   
   \item Second, if the complexity is taken into account, it should
   be said that the computation of $\equiv$ and $\sqsubseteq$
   for $(O,P,I)$
   requires the same time as entire clarification and reduction
   using classic methods described in basic literature \citep{GaWi:FCA}.
   Mao does not mention the complexity of the computation of $\equiv$ and $\sqsubseteq$
   neither does she refer to the classic methods.
\end{itemize}

Substitution of the basic notions with the equivalent newly introduced notions is Mao's main resource for results.
Instead of removal attributes
from $(O,P,I)$
she removes the corresponding vertices from $G(O,P,I)$.
Rephrased, using basic notions of FCA, the results become trivial or are already well-known.

Specifically, Theorem~3.2 states that duplicate attributes are reducible, 
which is obvious.
Theorem~3.3 states that in the clarified context $(O_2,P_2,I_2)$ no attributes are 
relatively necessary, which is obvious because $(O_2,P_2,I_2)$ is obtained
by clarification of the context $(O_1,P_1,I_1)$.
Theorem 3.4 is an overcomplicated characterization of reducible attributes, which after a straightforward simplification
becomes the one by \cite{GaWi:FCA}.
Theorem 3.5 states, that in the reduced context $(O_3,P_3,I_3)$ all attributes are irreducible,
which again is obvious, because $(O_3,P_3,I_3)$ is obtained 
by reduction of $(O_2,P_2,I_2)$.
Finally, Theorems~3.6 and 3.7 explain trivial relationships between Pawlak's types of attributes in 
$P_0,P_1,P_2$, and $P_3$.

\section{The proposed algorithms are incorrect}
\label{sec:alg}

 \cite{Mao2016} proposes a method of computing formal concepts in $(O,P,I)$; more specifically, 
the subset $\cal A$ containing all formal concepts excluding the top and the bottom concepts of $\B(O,P,I)$
and the attribute concepts.

\begin{figure}
\begin{center}
\begin{tikzpicture}

\begin{scope}[yscale=-1.5,xscale=2,yshift=-7cm]
\clip (0,0) to [out=30,in=-30](0,3) -- (0,3) to [out=-150,in=150] (0,0); 

\path[inner sep=0,yshift=1cm,draw,dashed,fill=gray!40] (0,0) to [out=45,in=-45](0,3) -- (0,3) to [out=-135,in=135] (0,0); 

\path[inner sep=0,yshift=2cm,xshift=-.2cm,fill=white] (0,0) to [out=45,in=-45](0,2) -- (0,2) to [out=-135,in=135] (0,0); 
\path[inner sep=0,yshift=2cm,xshift=.2cm,draw,dashed,fill=white] (0,0) to [out=45,in=-45](0,2) -- (0,2) to [out=-135,in=135] (0,0); 
\path[inner sep=0,yshift=2cm,xshift=-.2cm,draw,dashed] (0,0) to [out=45,in=-45](0,2) -- (0,2) to [out=-135,in=135] (0,0); 

\path[draw] (0,0) to [out=30,in=-30](0,3) -- (0,3) to [out=-150,in=150] (0,0); 

\node[inner sep=0,label={above:$c_1$}] at (0,1) {$\bullet$};
\node[inner sep=0,label={above:$b_1$}] at (-.2,2) {$\bullet$};
\node[inner sep=0,label={above:$b_2$}] at (.2,2) {$\bullet$};

\end{scope}

\begin{scope}[yscale=-1.5,xscale=2, yshift = -3.5cm]
\clip (0,0) to [out=30,in=-30](0,3) -- (0,3) to [out=-150,in=150] (0,0); 

\path[inner sep=0,yshift=1cm,draw,dashed] (0,0) to [out=45,in=-45](0,3) -- (0,3) to [out=-135,in=135] (0,0); 

\path[inner sep=0,yshift=2cm,xshift=-.2cm,fill=gray!40] (0,0) to [out=45,in=-45](0,2) -- (0,2) to [out=-135,in=135] (0,0); 
\path[inner sep=0,yshift=2cm,xshift=.2cm,draw,dashed,fill=gray!40] (0,0) to [out=45,in=-45](0,2) -- (0,2) to [out=-135,in=135] (0,0); 
\path[inner sep=0,yshift=2cm,xshift=-.2cm,draw,dashed] (0,0) to [out=45,in=-45](0,2) -- (0,2) to [out=-135,in=135] (0,0); 

\path[draw] (0,0) to [out=30,in=-30](0,3) -- (0,3) to [out=-150,in=150] (0,0); 

\node[inner sep=0,label={above:$c_1$}] at (0,1) {$\bullet$};
\node[inner sep=0,label={above:$b_1$}] at (-.2,2) {$\bullet$};
\node[inner sep=0,label={above:$b_2$}] at (.2,2) {$\bullet$};
\end{scope}

\begin{scope}[yscale=-1.5,xscale=2, yshift = 0cm]
\clip (0,0) to [out=30,in=-30](0,3) -- (0,3) to [out=-150,in=150] (0,0);

\path[fill=gray!40] (0,0) to [out=30,in=-30](0,3) -- (0,3) to [out=-150,in=150] (0,0);

\path[inner sep=0,yshift=1cm,draw,dashed,fill=white] (0,0) to [out=45,in=-45](0,3) -- (0,3) to [out=-135,in=135] (0,0); 

\path[inner sep=0,yshift=2cm,xshift=-.2cm,fill=white] (0,0) to [out=45,in=-45](0,2) -- (0,2) to [out=-135,in=135] (0,0); 
\path[inner sep=0,yshift=2cm,xshift=.2cm,draw,dashed,fill=white] (0,0) to [out=45,in=-45](0,2) -- (0,2) to [out=-135,in=135] (0,0); 
\path[inner sep=0,yshift=2cm,xshift=-.2cm,draw,dashed] (0,0) to [out=45,in=-45](0,2) -- (0,2) to [out=-135,in=135] (0,0); 

\path[draw] (0,0) to [out=30,in=-30](0,3) -- (0,3) to [out=-150,in=150] (0,0); 

\node[inner sep=0,label={above:$c_1$}] at (0,1) {$\bullet$};
\node[inner sep=0,label={above:$b_1$}] at (-.2,2) {$\bullet$};
\node[inner sep=0,label={above:$b_2$}] at (.2,2) {$\bullet$};
\end{scope}
\end{tikzpicture}
\end{center}
\caption{Portions of concept lattice computed by the algorithm~1\,(top), algorithm~2\,(middle), and algorithm~3\,(bottom)\label{fig:portions}}
\end{figure}
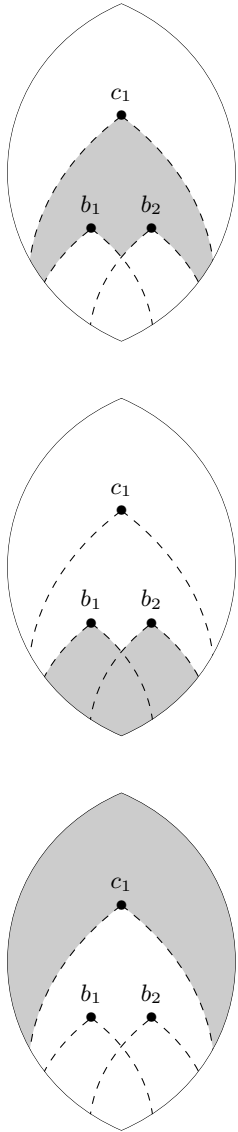

The method consists of three algorithms which each compute different portions of $\cal A$ with the aid of a pre-weighted relevant graph;
their input is a pre-weighted relevant graph and a maximal attribute $c_1$.
\begin{enumerate}
\item
The algorithm~1 computes the set ${\cal F}$ containing those concepts $\tu{A,B}\in {\cal A}$ which satisfy
$c_1 \in B$ and $b \notin B$  for all $b \in N^+_{G(O,P,I)}(c_1)$ (see Fig.\,\ref{fig:portions}\,(top)).
\item
The algorithm~2 computes the set ${\cal S}$ containing those concepts $\tu{A,B}\in {\cal A}$ which satisfy
$c_1 \in B$ and $b \in B$  for some $b \in N^+_{G(O,P,I)}(c_1)$ (see Fig.\,\ref{fig:portions}\,(middle)).
\item
The algorithm~3 computes the set ${\cal T}$ containing those concepts $\tu{A,B}\in {\cal A}$ which satisfy $c_1 \notin B$
(see Fig.\,\ref{fig:portions} (bottom)).
\end{enumerate}
$N^+_{G(O,P,I)}(c_1)$ in items 1. and 2. denotes the lower cone of attribute $c_1$ w.r.t. $\sqsubseteq$, excluding $c_1$ itself.

All three algorithms are described in a very complicated way which makes them almost unreadable.
More importantly, they are incorrect. In what follows, we present the examples in which the algorithms fail to deliver
correct outputs.

We need to recall some additional notation from \citep{Mao2016}:
\begin{itemize}
\item $V^{\to a}$ denotes the upper cone of attribute $a$ w.r.t. $\sqsubseteq$, excluding $a$ itself; 
\item $\omega(a)$ denotes the pre-weight of $a$, i.e. $\omega(a)=a^\down$.
\end{itemize}

\subsection{Example for algorithm 1}

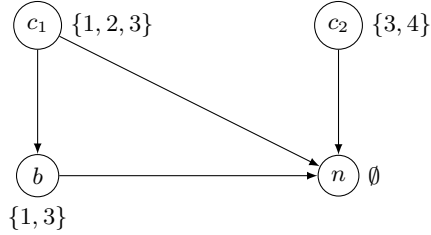
\begin{figure}[h]
\begin{center}
\begin{tikzpicture}
\begin{scope}[yshift=2.5cm,xshift=2cm]
\node (x) at (0,0) {%
\begin{tabular}{l|cccc}
  &$c_1$ &$c_2$&$b$ & $n$ \\\hline
1 &\X    &     &    &     \\
2 &\X    &     & \X &     \\
3 &\X    & \X  & \X &     \\
4 &      & \X  &    &     \\
\end{tabular}};
\end{scope}
\begin{scope}
\node[circle,draw,label={right:$\{1,2,3\}$}] (c1) at (0,0) {$c_1$};
\node[circle,draw,label={right:$\{3,4\}$}] (c2) at (4,0) {$c_2$};

\node[circle,draw,label={below:$\{1,3\}$}] (b) at (0,-2) {$b$};
\node[circle,draw,label={right:$\emptyset$}] (n) at (4,-2) {$n$};

\draw[-latex] (c1) -- (b);
\draw[-latex] (c1) -- (n);
\draw[-latex] (c2) -- (n);
\draw[-latex] (b) -- (n);
\end{scope}
\end{tikzpicture}
\end{center}
\caption{A formal context (top) and its pre-weighted relevant graph (bottom) as an example for algorithm 1\label{fig:cex1}}
\end{figure}

Consider the formal context depicted in Fig.\,\ref{fig:cex1}\,(top).

We demonstrate that the algorithm 1 fails to deliver correct output for its pre-weighted relevant graph (Fig.\,\ref{fig:cex1}\,(bottom)) and attribute $c_1$. We have ${\cal C} = \{ c_1, c_2 \}$ and $N^+_{G(O,P,I)(c_1)} = \{ b, n \}$. The following represents how the algorithm runs when we exactly follow exactly the steps in \cite{Mao2016}.

\medskip

\noindent
{\bf (step 1)}
\begin{align*}
H_1 = \{ x \in P \setminus \{c_1\} \mid &x \notin N^+_{G(O,P,I)}(c_1)\\
&\text{ and } \omega(c_1) \cap \omega(x) \neq \emptyset \} = \{ c_2 \}.
\end{align*}
As $H_1 \neq \emptyset$ is the case, we select $h_1 \in H_1$. There is only one option, thus we set $h_1 := c_2$.
We compute extent $A_1$ and $B_1$ as
\[
A_1 = \omega(c_1) \cap \omega(c_2) = \{ 3 \}  \quad\text{and}\quad B_1 = \{ c_1,c_2 \}.
\]
As $H_1 \setminus B_1 = \emptyset$, the condition of non-existence of $h \in H_1 \setminus B_1$ with $\omega(c_1)\cap \omega(c_2) \subset \omega(h)$ is trivially satisfied, and $\tu{A_1,B_1}$ is outputted.

\medskip

\noindent
{\bf (step 2)}
As $|H_1|=1$ we stop the computation.

We obtained a pair $\tu{\{3\},\{c_1,c_2\}}$ as an output, but this pair is not a formal concept. 
If we close it, we obtain $\tu{\{3\},\{c_1,c_2,b\}}$ which should not be in $\cal F$, since it contains $b$.

\subsection{Example for algorithm 2}

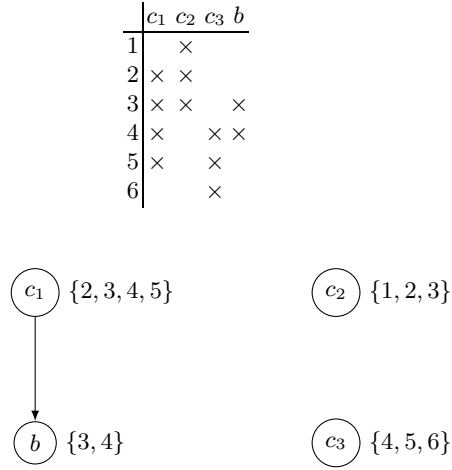
\begin{figure}[h]
\begin{center}
\begin{tikzpicture}
\begin{scope}[yshift=2.5cm,xshift=2cm]
\node (x) at (0,0) {%
\begin{tabular}{l|cccc}
  &$c_1$ &$c_2$ &$c_3$ &$b$ \\\hline
1 &      &  \X  &      &    \\
2 & \X   &  \X  &      &    \\
3 & \X   &  \X  &      & \X \\
4 & \X   &      & \X   & \X \\
5 & \X   &      & \X   &    \\
6 &      &      & \X   &    \\
\end{tabular}};
\end{scope}
\begin{scope}
\node[circle,draw,label={right:$\{2,3,4,5\}$}] (c1) at (0,0) {$c_1$};
\node[circle,draw,label={right:$\{1,2,3\}$}] (c2) at (4,0) {$c_2$};

\node[circle,draw,label={right:$\{4,5,6\}$}] (c3) at (4,-2) {$c_3$};
\node[circle,draw,label={right:$\{3,4\}$}] (b) at (0,-2) {$b$};

\draw[-latex] (c1) -- (b);
\end{scope}
\end{tikzpicture}
\end{center}
\caption{A formal context (top) and its pre-weighted relevant graph (bottom) as an example for algorithm 2\label{fig:cex2}}
\end{figure}

Consider the formal context depicted in Fig.\,\ref{fig:cex2}\,(top).

We demonstrate that algorithm 2 fails to deliver the correct output for its pre-weighted relevant graph (Fig.\,\ref{fig:cex2}\,(bottom)) and attribute $c_1$.
We have that ${\cal C} = \{ c_1, c_2, c_3 \}$ and $N^+_{G(O,P,I)}(c_1) = \{ b \}$.

\noindent
{\bf(step 6)} We set $b_1 := b$ as it is the only option; and we compute
\[
H_{b_1} = \{
c_2, c_3
\}.
\]

\noindent
{\bf(step 7)}
We have $H_{b_1} \neq \emptyset$. We select $d_1 := c_2$. We compute
\begin{align*}
A_{b_1} :&= \omega(b_1) \cap \omega(d_1)= \omega(b) \cap \omega(c_2) = \{ 3 \}, \\
B_{b_1} :&= \{ b_1,d_1 \} 
\cup V_3^{\to b_1}
\cup V_3^{\to d_1} = \{b,c_2\} \cup \{c_1\} \cup \emptyset\\ &= \{b,c_1,c_2\}.
\end{align*}
We continue with Case 2 because $c_3$, the only element in $H_{b_1} \setminus B_{b_1}$,
does not satisfy $A_{b_1} \subset \omega(c_3)$.
The rest of this step is trivial since $N^+_{G(O,P,I)}(b) = \emptyset$;
$\tu{A_{b_1}, B_{b_1}}$ is outputted.

\medskip

\noindent
{\bf(step 8)}
We set $d_{11} = c_2$, $d_{12} = c_3$. And compute 
\[
A_{1d_1} = \omega(b_1) \cap \omega(d_{11}) \cap \omega(d_{12}) = \omega(b) \cap \omega(c_2) \cap \omega(c_3) = \emptyset.
\]
Since $A_{1d_1}$ is empty the algorithm stops.

\medskip

The algorithm terminated without outputting a formal concept $\tu{ \{4 \}, \{c_1,c_3,b \} }$, which belongs to $\cal S$.

\subsection{Example for algorithm 3}

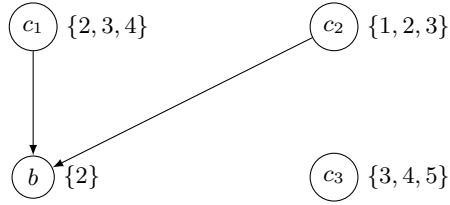
\begin{figure}[h]
\begin{center}
\begin{tikzpicture}
\begin{scope}[yshift=2.5cm,xshift=2cm]
\node (x) at (0,0) {%
\begin{tabular}{l|cccc}
  &$c_1$ &$c_2$ &$c_3$ &$b$ \\\hline
1 &      &  \X  &      &    \\
2 & \X   &  \X  &      & \X \\
3 & \X   &  \X  & \X   &    \\
4 & \X   &      & \X   &    \\
5 &      &      & \X   &    \\
\end{tabular}};
\end{scope}
\begin{scope}
\node[circle,draw,label={right:$\{2,3,4\}$}] (c1) at (0,0) {$c_1$};
\node[circle,draw,label={right:$\{1,2,3\}$}] (c2) at (4,0) {$c_2$};
\node[circle,draw,label={right:$\{3,4,5\}$}] (c3) at (4,-2) {$c_3$};
\node[circle,draw,label={right:$\{2\}$}] (b) at (0,-2) {$b$};

\draw[-latex] (c1) -- (b);
\draw[-latex] (c2) -- (b);
\end{scope}
\end{tikzpicture}
\end{center}
\caption{A formal context (top) and its pre-weighted relevant graph (bottom) as an example for algorithm 3\label{fig:cex3}}
\end{figure}

The third algorithm uses the previous two algorithms. Even if  algorithm~1 and algorithm~2 were correct,  algorithm 3 would be still incorrect, and this is shown for the formal context depicted in Fig.\,\ref{fig:cex3}\,(top), its pre-weighted relevant graph is in Fig.\,\ref{fig:cex3}\,(bottom), and its attribute $c_1$.

\medskip

\noindent
{\bf(step 15)}
${\cal C} := {\cal C} \setminus \{c_1\} = \{c_2,c_3\}$; 
\begin{equation}
\label{eq:upgradegopi}
G(O,P,I) := G(O,P,I) \setminus (c_1 \cup N^+_{G(O,P,I)}(c_1));
\end{equation}
 i.e. 
$G(O,P,I)$ now contains only vertices $c_2$ and $c_3$.

\medskip

\noindent
{\bf(step 16)} 
We use algorithms 1 and 2 for updated graph $G(O,P,I)$ and $c_2$.
It is not clear whether \eqref{eq:upgradegopi} includes removal of $c_1$ from $P$:
\begin{itemize}
\item
If yes, we get $\tu{\{3\},\{c_2,c_3\}}$ as one of the outputs of algorithm 1 and 2, but it is not a formal concept
\item
If no, we get $\tu{\{3\},\{c_1,c_2,c_3\}}$ as one of the outputs of algorithm 1 and 2, but it does not belong to $\cal T$.
\end{itemize}
Either way, we obtain a wrong output. 


\subsection{Bogus complexity analysis}

As the method of formal concept enumeration is 
incorrect, we could simply disregard 
its complexity analysis,
however, even then there is a problem 
which needs mentioning.

In the description of both, 
attribute reduction and formal concept enumeration,
Mao uses steps which perform intersections
and/or comparisons of pre-weights, i.e.
subsets of the set $O$. In the accompanied 
complexity analysis, she claims 
complexity of these steps to be ${\cal O}(|P|^2)$ or ${\cal O}(|P|^3)$; i.e 
independent of the size of the subsets
or the size of $O$.
This could be actually achieved if all the intersections of pre-weights
were precomputed,
however, a collection of all the intersections is, in fact, the collection of all extents.
As each extent uniquely determines its formal concept, this assumption means that Mao's
method of formal concept enumeration
requires all formal concepts to be precomputed.

To sum up, either the claimed complexities are
incorrect, or Mao works from an assumption that
makes the actual computation superfluous.

\section{Conclusion}


\cite{Mao2016} provides nothing new. Her theoretical results are either trivial or well-known and all three proposed algorithms are incorrect.

\section*{Acknowledgments}
Supported by grant No. 15-17899S, ``Decompositions of Matrices with Boolean and Ordinal Data: Theory and Algorithms'', of the Czech Science Foundation.

%

\end{document}